\documentclass{article}

\usepackage{microtype}
\usepackage{graphicx}
\usepackage{subfigure}
\usepackage{booktabs} 
\usepackage{soul}
\usepackage{url}

\usepackage{amsmath}
\usepackage{amssymb}
\DeclareMathOperator*{\argmax}{arg\,max}

\usepackage[accepted]{icml2019}

\newcommand{\YOON}[1] {
	\textcolor{red}{\bfseries{YOON: {#1}}}
}
\newcommand{\resp}[1] {
	\textcolor{blue}{\bfseries{resp: {#1}}}
}

\newcommand{\Skip}[1] {
}

\icmltitlerunning{Submission and Formatting Instructions for ICML 2019}

\begin{document}

\twocolumn[
\icmltitle{Intrinsic Motivation Driven Intuitive Physics Learning using Deep Reinforcement Learning with Intrinsic Reward Normalization}



\icmlsetsymbol{equal}{*}

\begin{icmlauthorlist}
\icmlauthor{Jae Won Choi}{to}
\icmlauthor{Sung-eui Yoon}{goo}
\end{icmlauthorlist}

\icmlaffiliation{to}{Department of Robotics, KAIST, Rep. of Korea}
\icmlaffiliation{goo}{School of Computing, KAIST, Rep. of Korea}

\icmlcorrespondingauthor{}{jaewonch@kaist.ac.kr}
\icmlcorrespondingauthor{}{sungeui@kaist.edu}

\icmlkeywords{Machine Learning, ICML}

\vskip 0.3in
]



\printAffiliationsAndNotice{}  

\begin{abstract}
At an early age, human infants are able to learn and build a model of the world
very quickly by constantly observing and interacting with objects around them.
One of the most fundamental intuitions human infants acquire is intuitive
physics. Human infants learn and develop these models, which later serve as
prior knowledge for further learning. Inspired by such behaviors exhibited by
human infants, we introduce a graphical physics network integrated with deep
reinforcement learning.
Specifically, we introduce an intrinsic reward
normalization method that allows our agent to efficiently choose actions that
can improve its intuitive physics model the most. 
 Using a 3D physics engine, we show that our
graphical physics network is able to infer object's positions and velocities
very effectively, and our deep reinforcement learning network encourages an
agent to improve its model by making it continuously interact with objects only
using intrinsic motivation. We experiment our model in both stationary and
non-stationary state problems and show benefits of our approach in terms of
the number of different actions the agent performs and the accuracy of agent's
intuition model. 

\end{abstract}

\section{Introduction}
Various studies in human cognitive science have shown that humans rely
extensively on prior knowledge when making decisions. Reinforcement learning
agents require hundreds of hours of training to achieve human level performance
in ATARI games, but human players only take couple hours to learn and play them
at a competent level. This observation begs the question, what prior knowledge
do humans have that accelerates their learning process? Recent works~\cite{lake_ullman_tenenbaum_gershman_2017} suggest that there are two core
ingredients which they call the `start-up software': intuitive physics and
intuitive psychology. These intuition models are present at a very early stage
of human development and serve as core knowledge for future learning. But how
do human infants build intuition models?

\begin{figure}[t]
\vskip 0.2in
\begin{center}
\includegraphics[width=\columnwidth]{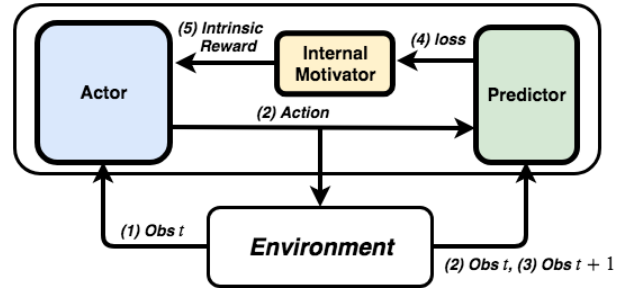}
\caption{This diagram illustrates how an agent chooses an action based on loss
incurred from its predictor which tries to mimic the behavior of an 
environment (or a subset of an environment). (1) After
making an observation at time $t$, the agent's actor module chooses
an action. (2) The predictor module takes the action and the observation at time
$t$ and makes a
prediction; (3) the predictor module then compares its prediction to the observation
from the environment at $t+1$ and (4) outputs a loss. (5) The loss value is 
converted into an intrinsic reward by the internal motivator module and is sent
to the actor's replay buffer to be stored for future
training. 
	}
\label{network}
\end{center}
\vskip -0.2in
\end{figure}

Imagine human infants in a room with toys lying around at a reachable distance. They are constantly grabbing, throwing and performing actions on objects; sometimes they observe the aftermath of their actions, but sometimes they lose interest and move on to a different object. The ``child as scientist" view suggests that human infants are intrinsically motivated to conduct their own experiments, discover more information, and eventually learn to distinguish different objects and create richer internal representations of them \cite{lake_ullman_tenenbaum_gershman_2017}. Furthermore, when human infants observe an outcome inconsistent with their expectation, they are often surprised which is apparent from their heightened attention \cite{acquisition_of_physical_knowledge}. The mismatch between expectation and actual outcome (also known as expectancy violations) have shown to encourage young children to conduct more experiments and seek more information~\cite{Stahl91, STAHL20171}.
\par

Inspired by such behaviors, in this paper, we explore intrinsically motivated
intuition model learning that uses loss signals from an agent's intuition model
to encourage an agent to perform actions that will improve its intuition model
(illustrated in Figure \ref{network}). Our contribution in this paper is
twofold: (1) we propose a graphical physics network that can extract physical
relations between objects and predict their physical behaviors in a 3D
environment, and (2) we integrate the graphical physics network with the deep
reinforcement learning network where we introduce the intrinsic reward
normalization method that encourages the agent to efficiently explore actions
and expedite the improvement of its intuition model.  The results show that our
actor network is able to perform a wide set of different  actions and our
prediction network is able to predict object's change in velocity and position
very accurately. 

\section{Approach}
In this section, we explain representations of objects and their relations, followed by our internal motivator.

\subsection{Representation}

We represent a collection of objects as a graph where each node denotes an
object and each edge represents a pairwise relation of objects. Due to its simple structure, we use sphere as our primary object throughout this paper. Given $N$ different objects, each object's state, $s_{obj_i}$, consists of its features, $d_{obj_i}$, and its relations with other objects:
\begin{align}
d_{obj_i} &= [x,y,z,vx,vy,vz,r,m] \,, \label{eq1}\\
r_{ij} &= f_r(d_{obj_i}, d_{obj_j}) \,, \label{eq2} \\
\mathbf{R}_{obj_i} &= \begin{bmatrix} r_{i1} & r_{i2} & \cdots & r_{iN}  \end{bmatrix}^T \,,  \label{eq3} \\
e_{obj_i} &=  \sum_{i \ne j, j\in [1,N]} r_{ij} \,, \label{eq4} \\
s_{obj_i} &= [d_{obj_i}, e_{obj_i}]  \,, \label{eq5}
\end{align}
where $d_{obj_i}$ contains object's Euclidean position $(x,y,z)$, Euclidean velocity
$(vx,vy,vz)$, size $r$ and mass $m$. Some of these features are not readily
available upon observation, and some works have addressed this issue by using
convolutional neural networks~\cite{blocktowers, visual_interaction_networks,
Fragkiadaki_billiards}; 
in
this paper, we assume that these features are known for simplicity.
To provide location invariance, we use an object-centric coordinate system, where the xy origin is the center of the sphere's initial position and the z origin is the object's initial distance to a surface (e.g. floor).
\par


\paragraph{Relation Encoder}
An object's state also includes its pairwise relations to other objects.
A relation from object $i$ to object $j$ is denoted as $r_{ij}$ and can be
computed from the relation encoder, $f_r$, that takes two different object
features, $d_{obj_i}$ and $d_{obj_j}$, and extracts their relation $r_{ij}$
(shown in Eq. (\ref{eq2})). 
Note that $r_{ij}$ is directional, 
and $r_{ij} \ne -r_{ji}$. Once all of object's pairwise relations are extracted, they are  stored in a matrix, $\mathbf{R}_{obj_i}$. The sum of all pairwise relations of an object yields $e_{obj_i}$, and when concatenated with the object feature
$d_{obj_i}$, we get the state of the object $s_{obj_i}$. 
\par
An observation is a collection of every object's state and its pairwise relations:
$$ obs = [(s_{obj_1}, \mathbf{R}_{obj_1}), (s_{obj_2}, \mathbf{R}_{obj_2})), \cdots , (s_{obj_n}, \mathbf{R}_{obj_n})] \,.$$

\subsection{Object-based Attention}
When an agent performs an action on an object, two things happen: (1) the
object on which action was performed moves and (2) some other object (or
objects) reacts to the moving object. Using the terms defined by
\cite{chang2016compositional},
we call the first object \textbf{focus object} and the other object
\textbf{relation object}. Given the focus object and the relation object, we 
associate agent's action to the observation of both focus object and relation object.  
The agent's job is to monitor the physical behaviors of both objects when selected action is performed and check whether its action elicited intended consequences. 

\subsection{Internal Motivator}

ATARI games and other reinforcement learning tasks often have clear extrinsic
rewards from the environment, and the objective of reinforcement learning is to
maximize the extrinsic reward. In contrast, intuition model learning does not
have any tangible reward from the external environment. For example, human
infants are not extrinsically rewarded by the environment for predicting how
objects move in the physical world.

In order to motivate our agent to continuously
improve its model, we introduce an internal motivator module. Similar to
the work of \cite{pathakICMl17curiosity}, the loss value from an intuition
model of an agent is converted into a reward by the internal motivator module
$\phi$: 
\begin{align}
r_t^{intrinsic} = \phi(loss_t) \label{eqphi}\,,
\end{align}
where $r_t^{intrinsic}$ is the
intrinsic reward at time $t$, and $loss_t$ is loss from a prediction network at
time $t$.
While not having any extrinsic reward may result in ad hoc
intrinsic reward conversion methods, we show that our model actually benefits
from 
our simple, yet effective intrinsic reward normalization method.
Note that while our method adopted to use the prediction error as the loss
based on a prior work~\cite{pathakICMl17curiosity, pathak18largescale}, our approach mainly focuses on
constructing an intuitive physics model, while the prior approaches are designed
for the curiosity-driven
exploration in a game environment.


\textbf{Intrinsic Reward Normalization}
In conventional reinforcement learning
problems, the goal of an agent is to maximize the total reward. To do so, it
greedily picks an action with the greatest Q value. If that action does indeed
maximize the overall reward, the Q value of that action will continue to
increase until convergence. However, sometimes this could cause an agent to be
stuck in a sub-optimal policy in which the agent is oblivious to other actions
that could yield a greater total reward. 

To ameliorate the problem of having a sub-optimal policy, the most commonly
used method is a simple $\epsilon$-greedy method, which randomly chooses an
action with $\epsilon$ probability. This is a simple solution, but
resorts to randomly picking an action for exploration. 
An efficient way for
exploration in stationary state setting with discrete actions is to use the
upper confidence bound (UCB) action selection \cite{Sutton:1998:intro_to_rl}:
$$a_t = \argmax_a \Big[ Q_t(a) + c \sqrt{\frac{\ln{t}}{N_t(a)}} \Big] \,,$$
where $Q_t(a)$ is the Q value of action $a$ at time $t$, $N_t(a)$ is the number
of times action $a$ was chosen. However, this method has several shortcomings:
(1) it is difficult to be used in non-stationary states since UCB does not
take states into account, and (2) it can be more complex than $\epsilon$-greedy
method since it needs to keep track of the number of actions taken. Notice that
this heuristic assumes that Q value of some action $a$ at time $t$, or
$Q_t(a)$, converges to some upper bound $\mathcal{U} \in \mathbb{R}$; i.e., $Q_t(a)
\to \mathcal{U}^-$ as $t \to \infty$. 

Intuitive physics learning is, however, slightly different from that of
conventional reinforcement learning in that $\lim\limits_{t \to \infty} Q_t(a)
= 0$ $\ \forall a$. To show this, assuming that our intuition model improves in accuracy, it is equivalent to saying that loss is decreasing: $loss_t \to 0 \text{ as } t \to \infty \,.$
Earlier, we defined our intrinsic reward $r^{intrinsic}_t = \phi(loss_t)$ at Eq. (\ref{eqphi}). 
Since $loss_t \to 0$, we know that $r_t^{intrinsic} \to 0$ as long as $\phi$ is
a continuous and increasing function. 
In both stationary and non-stationary state problems, if we train $Q_t(a)$ to
be $r_{t}^{intrinsic}$, we can show that $\lim\limits_{t \to \infty} Q_t(a) =
0$.
\Skip{
 \st{However, if we train $Q_t(a)$ 
in an infinite state space, we cannot prove that $Q_t(a) \to 0$) due to the
complicated nature of asymptotic analysis} \YOON{Am I right?}. \resp{I'm not sure. This is a bit more complex. I'd prefer just not mentioning it for now and come back later }
}
\par

In order to capitalize on this property, we first normalize $\phi(loss_t)$ by some upper bound $\mathcal{U}$ to get a normalized reward:
\begin{equation} \label{eq:normR}
R^{intrinsic}_t = \frac{\phi(loss)}{\mathcal{U}}\,,
\end{equation} 
which restricts $R^{intrinsic}_t \in [0,1]$. Additionally, we normalize all
$Q_t(a)$ by using the following equation, which is also known as the Boltzmann
distribution of actions \cite{Sutton:1998:intro_to_rl}: 
\begin{align}
\mathcal{Q}_t(a) &= \Big[ \frac{e^{Q_t(a)}}{\sum_{k=1}^{|A|} e^{Q_t(k)}}  \Big]\,, \label{eq:normQ} \\
A_t &=  \argmax_{a'} \mathcal{Q}_t(a') \label{eq:argmaxa}\,,
\end{align}
where $\mathcal{Q}_t(a)$ is the normalized Q value of action $a$ at time $t$,
$|A|$ is the cardinality of the discrete action space, and $A_t$ is the action
with the highest, normalized Q value.

Given $\mathcal{Q}_t(a) \in [0,1]$ and $R^{intrinsic}_t \in [0,1]$, as we train
$\mathcal{Q}_t(a)$ with $R^{intrinsic}_t$ using gradient descent, this will
naturally increase $Q_t(a)$ of actions that have not been taken according to
our normalization step (Eq.~\eqref{eq:normQ}); i.e., $Q_t(a')$ of  $a'$ actions
that have been taken decreases and thus other actions that were not taken will
have relatively bigger Q values.


\begin{figure*}[t]
\vskip 0.2in
\begin{center}
\includegraphics[width=\textwidth]{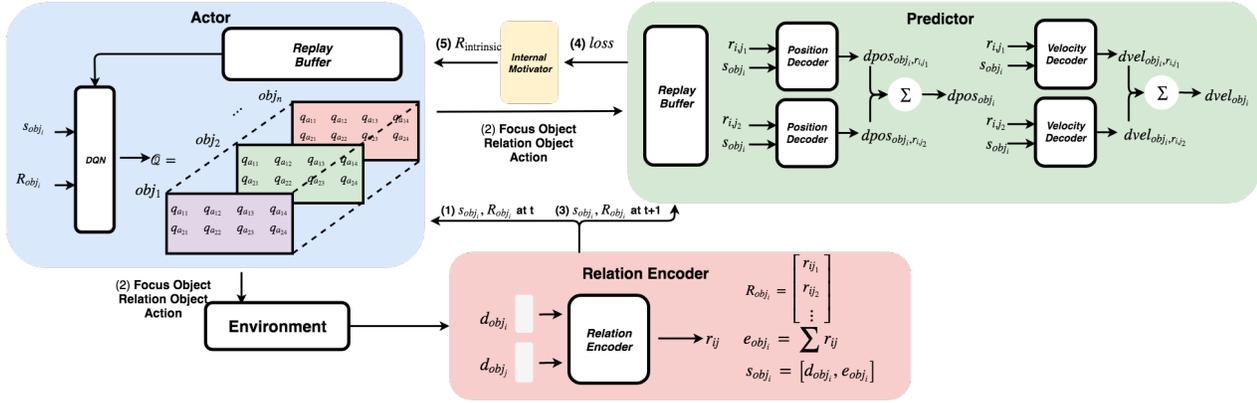}
\end{center}
\caption{
	This shows a detailed diagram of our overall approach shown in
	Fig.~\ref{network}.	(Bottom Right) For every pair of objects, we
	feed their features into our relation encoder to get relation $r_{ij}$
and object
	i's state $s_{obj_i}$. (Top Left) Using the greedy method, for each
	object, we find the maximum Q value to get our focus object, relation object, and action. (Top Right) Once we have our focus object and
	relation object, we feed their states and all of their relations into
	our decoders to predict the change in position and change in velocity.}
\label{network_detail}
\vskip -0.2in
\end{figure*}

Compared to other methods such as UCB, our method has the benefit of not
needing to explicitly keep track of the number of actions taken, and instead
takes advantage of the decreasing behavior of loss values to encourage
exploration. 
Moreover, because our method does not keep track of actions, it can be used in
non-stationary state problems as well, which we show later in 
Sec.~\ref{sec_exp}. 

\subsection{Replay Buffers}

Our agent has two separate replay buffers: one to store object's reward values
to train the actor (\textbf{actor replay buffer}) and another to store the physical
behaviors of objects to train the intuitive physics model (\textbf{prediction replay buffer}).
For both networks, past experiences are sampled uniformly despite the fact that
human brains are able to selectively retrieve past experiences from memory.

Despite uniform sampling, as our agent continuously experiments with objects,
both replay
buffers are more likely to contain experiences that the agent predicted with
low accuracy. 
This is because our agent will greedily choose an action with the greatest Q
value as shown in Eq.~\eqref{eq:argmaxa} and by design, action with the
greatest Q value also has the greatest loss (equivalently low accuracy) by
Eq.~\eqref{eq:normR}.  Nonetheless, if the replay buffer is not big enough,
this could let the agent overfit its intuitive physics model to experiences
with
high stochasticity. An ideal solution is to let agent curate its replay buffer
content to find a set of experiences that can maximize the generalizability of
the network. There are works that have addressed similar issues such as prioritized
replay buffer \cite{prioritizedRB}.
However, we use uniform sampling in our work for its simplicity.

\section{Network Models}

Using deep neural networks, our agent network can be separated into three
different sub-networks: relation encoders, deep Q network, and position \&
velocity decoders, illustrated in Fig.~\ref{network_detail}.

\subsection{Deep Q Network}
\label{subsec_dqn}

Inspired by the recent advances in deep reinforcement learning
~\cite{mnih-dqn-2015, Schulman:2015:TRP:3045118.3045319,
DBLP:journals/corr/SchulmanWDRK17,
Silver:2014:DPG:3044805.3044850}, we use the object oriented 
~\cite{Diuk_ORE}
deep Q network to choose three things: (1) focus object, (2) relation object,
and (3) action. 

For each object, our deep Q network takes $\mathbf{R}_{obj_i}$ and
$s_{obj_i}$ as input and computes $(N-1) \times |A|$ matrix, whose column
represents action Q values and row represents $obj_i$'s relation object.
Computing this for all objects, the final output of the network has a
shape of $N \times (N-1) \times |A|$, which we call $\mathcal{Q}$,
shown in the top left module 

in Fig.~\ref{network_detail}. 
Our agent greedily finds focus
object, relation object and action:
$$\textit{focus object, relation object, action} = \argmax_{i,r,a}
\mathcal{Q}_{i,r,a}\,,$$ 
where $i$ indicates the focus object index, $r$ is the relation object index, and $a$  denotes the action index.
\Skip{  
\YOON{conflicting.. Eq. 9 only choses the action. how do you compute other ones?It requires explanation..} \resp{fixed by not mentioning eq 9 since Eq. 9 is to show the concept of reward normalization. This is implementation detail.}
}

\par
Our deep Q network does not use a target network, and the actor replay buffer
samples experiences of every object, instead of randomly
sampling from all
experiences uniformly. This is done to prevent an agent from interacting with
only one object and from generalizing the behavior of one object to other
objects. 
\par

\par
With this setup, we set our stationary state target Q value, $y^{s}_t$, to be:
$$y^{s}_t =  R^t_{intrinsic} \,,$$ 
and non-stationary state target Q value, $y^{ns}_t$, to be
\begin{equation} \label{eqdqn}
y^{ns}_t =  \begin{cases} 
      R^{intrinsic}_t  \hspace{0.7cm}  \text{if reset occurs at t+1 } \\
      \min(1, (R^{intrinsic}_t + \gamma \max_{a'} Q_{a'}(s_{obj_i}^{t+1}))) \text{ o.w.} &
   \end{cases},
\end{equation}
where $\gamma$ is a discount factor in $[0,1]$ and objects' states reset when
one of the objects goes out of bounds. We provide details of non-stationary
state
experiment and bounds in Sec.~\ref{sec_exp_setup}. For stationary state
problems, since there is only single state,
we only use $R^{intrinsic}_t$ to
update our Q value. For non-stationary problems, we take the subsequent state
into account and update the Q value with the sum of reward and discounted next
state Q value. Note that our Q values and rewards reside in $[0,1]$ because of
the reward normalization method; therefore, when the sum of reward and
discounted next state reward exceeds 1, we clip the target value to 1.

\subsection{Position \& Velocity Decoders}
The predicted position and velocity of each object is estimated by the predictor module which is placed inside a green box in Fig.~\ref{network_detail}).
 
An object's state, $s_{obj_i}$ and all of its pairwise relations,
$\mathbf{R}_{obj_i}$, are fed into both position and velocity decoders to
predict the change in position and change in velocity of $obj_i$. For each
pairwise relation, $r_{ij}$, we get an output $dpos_{i,r_{ij}}$ from the position
decoder and $dvel_{i,r_{ij}}$ from the velocity decoder. 

Once all relations are accounted for, the sum of all $dpos_{i,r_{ij}}$ and
$dvel_{i,r_{ij}}$ are the final predicted change in position, $dpos_{obj_i}$,
and predicted change in
velocity, $dvel_{obj_i}$, of an object $i$:
\begin{align*}
    dpos_{obj_i} &= \sum_{i \ne j, j\in [1,N]} dpos_{obj_i, r_{i,j}}\,, \\
    dvel_{obj_i} &= \sum_{i \ne j, j\in [1,N]} dvel_{obj_i, r_{i,j}}\,,
\end{align*}
We train both decoders and relation encoder with the sum of mean squared errors
of positions and velocities:
\begin{align*}
loss = \sum_{k=\{i,r\}} & || dpos_{obj_k} - dpos'_{obj_k} ||_2 \\ & + || dvel_{obj_k} - dvel'_{obj_k} ||_2\,,
\end{align*}
where $i$ is the focus object index and $r$ is the relation object index.
$dpos'_{obj_k}$ and $dvel'_{obj_k}$ are the ground truth change in position and
velocity of an object, and are readily available 
by the physics engine.

\section{Experiment Setup}
\label{sec_exp_setup}

\begin{figure}[t]
\centering
\includegraphics[scale=0.1]{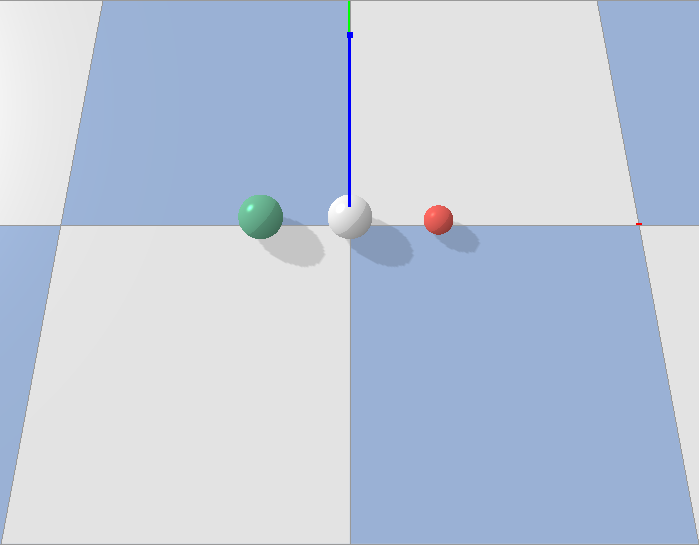}
\includegraphics[scale=0.1]{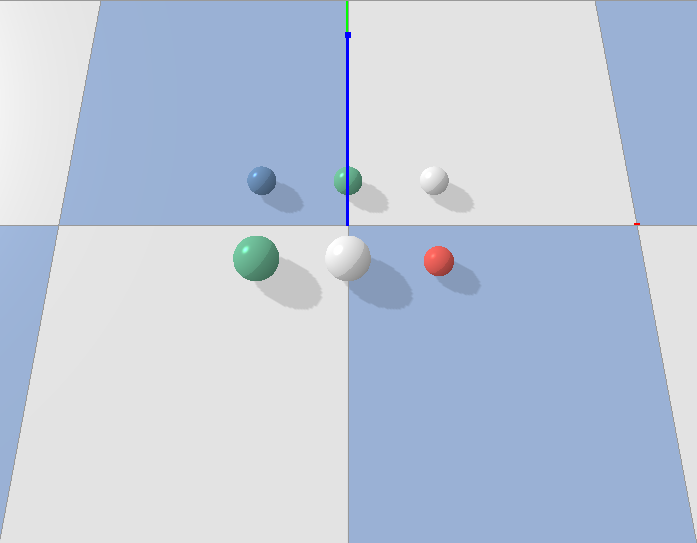}
\includegraphics[scale=0.1]{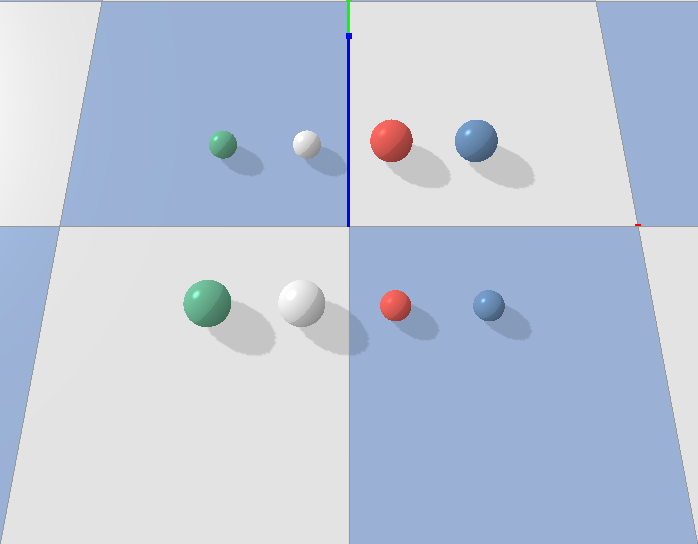}
\caption{Experiments in three different scenes: 3-object, 6-object and 8-object
	scenes. Colors represent different weights where red is 1kg, green is
	0.75kg, blue is 0.5kg and white is 0.25kg. The radius of each object
	can be either 5cm or 7.5cm. We experiment in two scenarios: stationary
	and non-stationary states. }
\label{experiment_setups}
\end{figure}

\textbf{Objects } In our experiment, we use spheres as primary objects due to
its simple structure (i.e. we can represent an object with only one variable -
radius). We use the center of the sphere as its xy position and its distance to
a surface, i.e. floor, as z position. 
We used pybullet as our physics simulator with timestep of 1/240 seconds with
30 fps.  As shown in Figure \ref{experiment_setups}, we use three different
scenes: 3-object, 6-object, and 8-object scenes. Objects are color coded so
that each denotes different weight: red is 1kg, green is 0.75kg, blue is 0.5kg,
and white is 0.25kg. Each object can have radius of 5cm or 7.5cm.

\textbf{Normalized Action Space } We provide an agent with actions in three
different directions: x, y and z. We experiment with a set of 27 actions
($x, y \in \{ -1.0, 0.0, 1.0 \}$ and $z \in \{ 0.0, 0.75, 1.0
\} $), and a set of 75 actions ($x, y \in \{ -1.0, -0.5, 0.0, 0.5, 1.0 \}$ and
$z \in \{ 0.0, 0.75, 1.0 \} $).
An action chosen from the Q network is then multiplied by the max
force, which we set to 400N.

\textbf{Performance Metric } Unlike conventional reinforcement learning
problems, our problem does not contain any explicit reward that we can maximize
on. For that reason, there is no clear way to measure the performance of our
work. Outputs from both prediction network and deep Q network rely heavily on
how many different actions the agent has performed. For instance, if an agent performs
only one action, the prediction loss will converge to 0 very quickly, yet agent would have only
learned to predict the outcome of one action. Therefore, we provide  the
following performance metrics to see how broadly our agent explores different
actions and how accurately it predicts outcomes of its actions:
\begin{table*}[t]
\caption{\textbf{Stationary state Action Coverage} We record the number of interactions our agent takes to perform actions in three different scenes. Each scene was experimented with two or more trials with different random seeds. We stop training when interaction count exceeds 75000 interactions. }
\label{stationary_state_result}
\vskip 0.15in
\begin{center}
\begin{small}
\begin{sc}
\begin{tabular}{lcccr}
\toprule
& \#Actions per relation & Total \# Actions & Action Coverage &  Interaction Count\\
\toprule
3-Object Scene  & 27 & 162& 1.0  & $770.38 \pm 187.914$ \\
& 75 & 450 & 1.0  & $2505.6 \pm 401.968$ \\
6-Object Scene  & 27 & 810 & 1.0  & $7311.0 \pm 2181.74$ \\
 & 75 & 2250 & 1.0  & $25958.67\pm 3609.25$ \\
8-Object Scene  & 27  & 1512 & 1.0 & $22836 \pm 2225.0 $ \\
& 75 & 4200 & $0.8 \pm 0.107$ & $75000$ \\
\bottomrule
\end{tabular}
\end{sc}
\end{small}
\end{center}
\vskip -0.1in
\end{table*}

\begin{itemize}
\item \textbf{Action Coverage } We measure the percentage of actions covered by
	an agent. 
	There are in total of $N \times (N-1) \times |A|$
		many actions for all pairs of objects.
        We use a binary matrix, $M$, to keep track
		of actions taken. We say that
		we covered a single action when
		an agent performs that action
		on all object relation pairs for every focus object. In short,
		if $\prod_{i,r} M_{i,r,a_k} = 1 \text{ for some action $a_k$}$,
		then we say it covered the action $a_k$. Action coverage is then
		computed by $(\sum_a \prod_{i,r} M_{i,r,a}) / |A|$. Action
		coverage value will tell us two things: (1) whether our
		predictor module is improving and (2) whether our agent is exploring
		efficiently. If the predictor module is not improving, the
		intrinsic reward associated with that action will not decrease,
		causing the agent to perform the same action repeatedly.

\item \textbf{Prediction Error } Once an agent has tried all actions, we use prediction error to test whether agent's predictor module improved.
\end{itemize}
For other hyperparameters, we set the upper bound $\mathcal{U}$ to be infinity and $\phi$ to be an identity function.

\section{Experiments}
\label{sec_exp}

In this section, we provide results of our experiments in 
stationary and non-stationary state problems. 

\subsection{Stationary State (Multi-armed Bandit Problem)}
In stationary state, or multi-arm bandit problem, after an agent takes an
action, we reset every object's position back to its original state. The
initial states of objects in different scenes are shown in Figure
\ref{experiment_setups}. We test with two different action sets and compare the
number of interactions it takes for an agent to try out all actions. We also
test generalizability of our agent's prediction model to multiple objects.
\par

\textbf{Action Coverage } 
As shown in Table \ref{stationary_state_result}, our agent
successfully tries all available actions in 3-object and 6-object scenes for both set of actions. Despite 
the huge number of actions, our agent is able to intrinsically motivate itself to perform hundreds and thousands of actions. 
While there is no clear way to tell our method is the fastest, 
we are not aware of any previous work that measures the number of actions
covered by an agent, since conventional reinforcement learning problems do not
require an agent to perform all actions. The fastest way to cover all actions
is to keep track of all actions and their Q values in a table; however, this
method has scalability issues and cannot be extended to non-stationary state
problems. 

As number of actions increases, it, however,  takes longer for our agent to
cover all actions. 
In fact, for 8-object scene, it fails to achieve full action coverage when
presented with 75 different actions per object. One possible explanation of
this is the replay buffer. Using a sampling batch size of 1024, the actor
replay buffer uniformly samples 1024 past experiences for all objects. For
8-object scene with 75 actions, each object has 525 unique actions 
per all pairs of objects. 
Compared to other scenes, the probability of getting 525 unique action
experiences from 1024 samples is a lot lower, especially when the actor replay
buffer has uneven distribution of actions. 

\begin{figure*}[t]
\includegraphics[width=\textwidth/3]{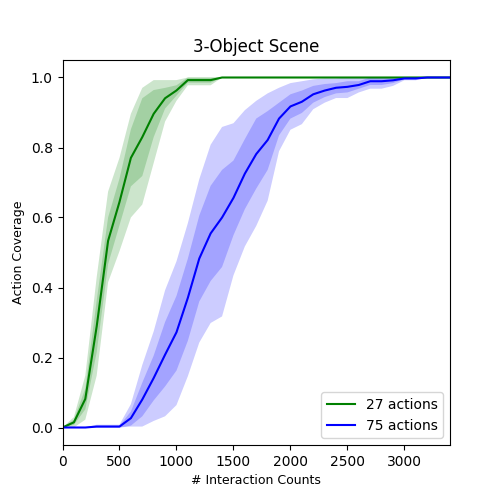} \includegraphics[width=\textwidth/3]{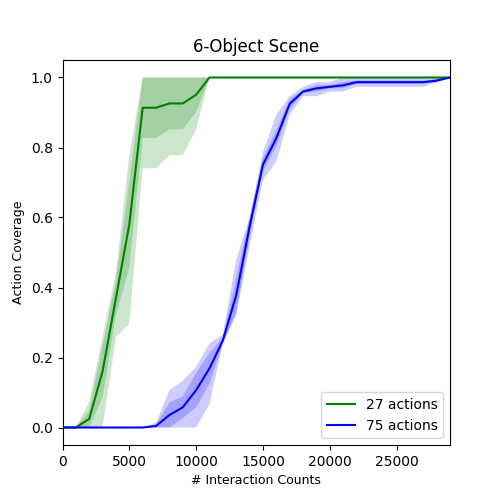} \includegraphics[width=\textwidth/3]{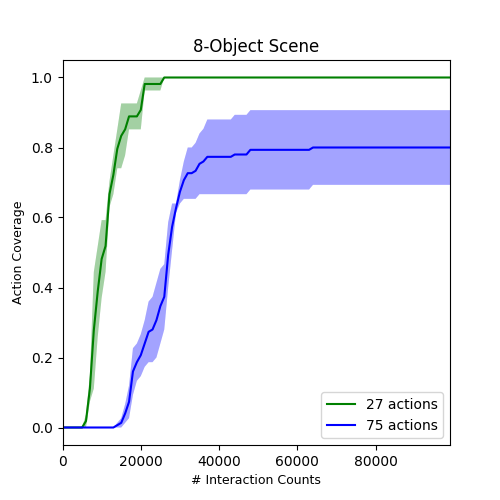}
	\caption{Action coverage results after 3 or more runs with different random seeds. (From left to right) Action coverage of our agent in 3-object, 6-object, and 8-object scene.
	}
\label{result_plots}
\end{figure*}

\Skip{
This problem is also noticeable in
3-object and 6-object scenes as shown in Figure \ref{result_plots}. The more
actions agent has taken, the longer it takes for the agent to find actions not
taken shown by the slow rate of convergence as it approaches to 1.
}

To make
matters worse, our prediction replay buffer is limited in size. In our experiment, the prediction
buffer of size $2.5e+6$ saturates 
before our
agent can perform all actions..  
It is apparent that our approach cannot scale
to scenes with more objects and actions with the current implementation of both
actor and prediction replay buffers. We leave the improvement of replay buffers
to future work.  \par

\Skip{
	\YOON{PUt this one into the sup. report, if necessary.}
\begin{figure}[t]
\begin{center}
\includegraphics[scale=0.5]{images/replay_buffer_tf_plot.png}
	\caption{Prediction replay buffer in 8 object scene with 75 actions}
\end{center}
\label{pred_replay_plot}
\end{figure}
}

\begin{figure}[tbh!]
\begin{center}
\includegraphics[scale=0.5]{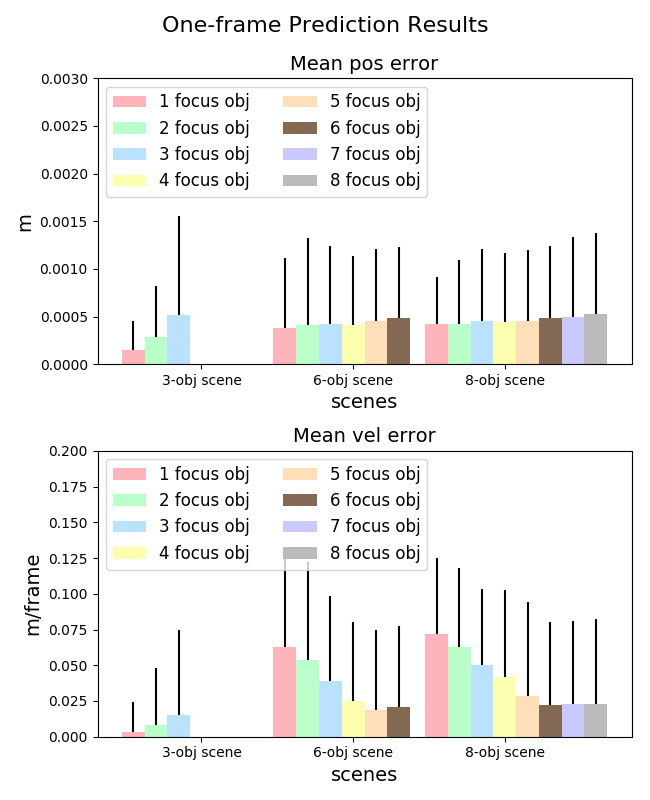}
\caption{Mean position and velocity prediction errors after 1 frame with different number of focus objects in 3-object, 6-object, and 8-object scene.
}
\label{losses}
\end{center}
\end{figure}

\textbf{Prediction Error } To test whether our agent's intuitive physics
model predicts object position and velocity correctly, we test it by computing
the $L2$-norm between predicted and the actual position and velocity of object
after one frame. Our agent's prediction model errors are plotted in Figure
\ref{losses}. For all scenes, the position errors of one frame prediction are
within 0.002m and the velocity errors are within 0.15m/s. These errors quite
small, given the fact that  the objects in our scene can change its position
from 0 to 0.18465m per frame and velocity change ranges from 0 to 6.87219m/s
per frame. 

Our prediction error could be reduced further with other network architectures.
In fact, there are many works that are trying to develop better network models
for intuitive physics learning. However, the aim of our work is to show that
the loss value from any intuitive physics network can be converted into an
intrinsic reward to motivate an agent to explore different actions.
Observations from different actions result in a diverse set of training data,
which can make the intuitive physics network more general and robust.

In generalization tasks, we allow our agent to apply forces
on multiple objects and see if it can predict the outcome after one frame. We generate 100 experiments by
randomly selecting focus objects and actions. We let our agent to predict the
next position and velocity of all objects in the scene, and we measure the mean
error of all predictions. Video of our results can be found in
supplementary video 1. 
The results show that our agent's prediction network accurately predicts the 
physical behavior of colliding and non-colliding objects. Moreover, our agent's intuition model is
able to generalize to multiple moving objects very well even though the agent
was only trained with an observation of a pair of objects (i.e.focus and relation object). 
Our qualitative results show that the agent's prediction model is able to 
predict that collision causes a moving object to either deflect or stop, and causes idle objects to move. 
Additionally, despite not knowing about gravitational forces, it learns that objects fall when they are above a surface. 

While there are other intuitive physics networks trained with supervised
learning that can yield a higher accuracy, it is difficult for us to compare
our results with theirs. The biggest reason is that supervised learning
requires
a well-defined set of training data a
priori. In our work and reinforcement learning in general, data are collected
by an agent, and those
data 
are not identical to that of supervised learning, making it difficult to
compare two different approaches in a fair setting. Additionally, the training
process differs: supervised learning takes epoch based learning where it
iterates over the same dataset multiple times until the network reaches a
certain error rate on a validation set. In deep reinforcement learning, a small
subset of data is
randomly sampled from the replay buffer and is used to train
the network on the fly. 

\subsection{Non-stationary State (Reinforcement Learning)} 
We extend our prediction model with deep Q network to non-stationary state
problems where we do not reset the objects unless they go out of bounds. To increase the chance of 
collision, we provide 9 actions only on the $xy$ place (i.e. $x,y \in \{-1, 0, 1\}$). We
arbitrarily set the bounds to be $3m \times 3m$ square. Since there are no
walls to stop objects from going out of bounds, 
objects have a low probability of colliding with another object. In order to make
them collide on every interaction, we generated 51 test cases in which every
interaction causes a collision among objects. Our agent's prediction model then
predicts the object's location after varying number of frames (i.e.
1,2,4,10,15,30,45 frames).  

\begin{figure}[tbh!]
\includegraphics[width=\columnwidth]{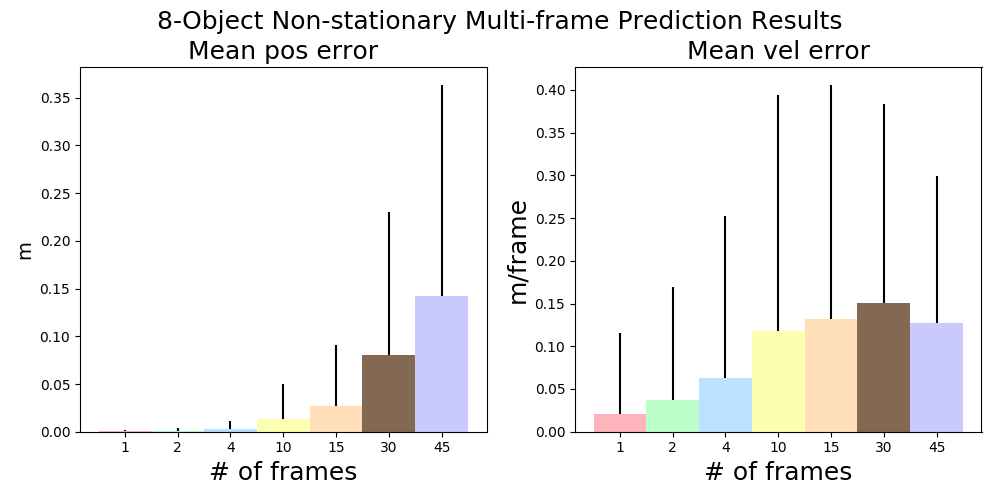} 
\caption{We test our intuition model in the non-stationary problem. Fewer
	prediction steps incur almost no error, while higher prediction steps,
	i.e. more than 4 frames, incur a high loss due to cumulative errors.}
\label{nonstationary_plots}
\end{figure}

\textbf{Prediction Error }
Prediction network's errors are plotted in Figure
\ref{nonstationary_plots}.
We see that when our agent predicts object locations after 1, 2, and 4 frames,
the error is negligibly small. However, our agent is uncertain when predicting
object location after 10 or more frames. This is because the error from each
frame accumulates and causes objects to veer away from the actual path. Our qualitative 
results of non-stationary problem can be found in 
supplementary video 2. 
Similar to stationary state results, our agent's prediction network 
accurately predicts an object's general direction and their movements, despite having infinitely many states. 
The agent's prediction network is able to predict whether an object will stop or deflect from its original trajectory 
when colliding with another object. Even if the agent's prediction network fails to predict the 
correct position of a moving object, it still makes a physically plausible prediction. 


\textbf{Limitation }
We would like to point out that although our agent performed thousands of
interactions, our agent still fails to learn that objects do not go through one
another, as seen in our qualitative results. This is very noticeable
when objects are moving fast. We conjecture that once our agent makes a wrong
prediction, the predicted object overlaps with another object, causing our
agent fails to predict subsequent positions and velocities. One plausible
explanation is that it has never seen such objects overlap in its training
data, hence fails to predict it accurately.
\Skip{
\YOON{we can make such penetrating case and a sequence of resolving such penetration in the simulation. This is one of critical issues developed in computer graphics simulation}. 
\st{It remains unclear how concepts such
as object solidity, contact, and other properties are learned by human infants}\YOON{point
that this capability is a start-up package and cite some package}, \st{and even
after thousands of interactions with objects, such concepts remain elusive to our agent} \resp{delete}.  
}

\section{Background}

\subsection{Deep Reinforcement Learning}
Recent advances in deep reinforcement
learning have achieved super-human performances on various ATARI games
\cite{mnih-dqn-2015} and robotic control problems
\cite{Schulman:2015:TRP:3045118.3045319, DBLP:journals/corr/SchulmanWDRK17,
Silver:2014:DPG:3044805.3044850}. While these approaches have achieved
state-of-the-art performances on many tasks, they are often not easily
transferable to other tasks because these networks are trained on individual
tasks. 

As opposed to model-free methods, model-based approaches create a model of an
environment, which equips agents with the ability to predict and plan.  Dyna-Q,
proposed by \cite{Sutton90integratedarchitectures}, integrated
model-free with model-based learning so an agent can construct a model of an
environment, react to the current state and plan actions by predicting future
states. More recent work in model based reinforcement learning~\cite{value_prediction_network} proposed a value prediction network that
learns a dynamic environment and predicts future values of abstract states
conditioned on options.

\subsection{Intrinsic Motivation and Curiosity}
Early work by \cite{Berlyne25_curiosity_and_exploration} showed that both
animals and humans spend a substantial amount of time exploring that is driven
by curiosity. Furthermore, Berlyne's theory suggests that curiosity, or
intrinsic motivation~\cite{intrinsically_motivated_hierarcial_skills,
intrinsic_rl}, is triggered by novelty and complexity. 

The idea of integrating
curiosity, and its counterpart boredom, with reinforcement learning was
suggested by \cite{Schmidhuber91apossibility_for_curiosity_boredom}, and showed
that intrinsic reward can be modeled to motivate agents to explore areas with
high prediction errors~\cite{Formal_theory_of_creativity_fun_im}. Using deep
learning, \cite{pathakICMl17curiosity} proposed an intrinsic curiosity module
that outputs a prediction error in the state feature space, which is used as an
intrinsic reward signal. Our work adopted this approach of using the prediction
error for our intrinsic reward. 

\subsection{Intuitive Physics}
At an early age, human infants are equipped with a ``starter pack" \cite{lake_ullman_tenenbaum_gershman_2017}, which includes a sense of intuitive physics. For instance, when observing a moving ball, our intuitive physics can sense how fast the ball is going and how far the ball will go before it comes to a complete halt. This intuitive physics is present as a prior model and accelerates future learning processes. Works done by \cite{Battaglia18327_simulation_as_an_engine} and \cite{internal_physics_models} show that humans have an internal physics model that can predict and influence their decision making process. 

Recent works have focused on using deep learning to model the human's intuitive
physics model. \cite{blocktowers} used a 3D game engine to simulate towers of
wooden blocks and introduced a novel network, PhysNet, that can predict whether
a block tower will collapse and its trajectories. \cite{Fragkiadaki_billiards}
proposed a visual predictive model of physics where an agent is able to predict
ball trajectories in billiards. Another work by \cite{chang2016compositional}
proposed a neural physics engine that uses an object based representation to
predict the state of the focus object given a physical scenario.
\cite{InteratcionNetwork} presented an interaction network that combined
structured models, simulation, and deep learning to extract relations among
objects and predict complex physical systems. We extend the previous works by 
integrating deep reinforcement learning that intrinsically motivates our agent to improve its physics model. 

\section{Discussion}
In this paper, we have proposed a graphical physics network integrated with
deep Q learning and a simple, yet effective reward normalization method that
motivates agents to explore actions that can improve its model. We have
demonstrated that our agent does indeed explore most of its  actions, and our
graphical physics network is able to efficiently predict object's position and
velocity. We have experimented our network on both stationary and
non-stationary problems in various scenes with spherical objects with varying
masses and radii.
Our hope is that these pre-trained
intuition models can later be used as a prior knowledge for other goal oriented
tasks such as ATARI games or video prediction.

\section*{Acknowledges}
This work was supported by  NRF/MSIT (No. 2019R1A2C3002833) and
IITP-2015-0-00199.

\bibliography{example_paper}

\begin{thebibliography}{27}
\providecommand{\natexlab}[1]{#1}
\providecommand{\url}[1]{\texttt{#1}}
\expandafter\ifx\csname urlstyle\endcsname\relax
  \providecommand{\doi}[1]{doi: #1}\else
  \providecommand{\doi}{doi: \begingroup \urlstyle{rm}\Url}\fi

\bibitem[Baillargeon(2007)]{acquisition_of_physical_knowledge}
Baillargeon, R.
\newblock \emph{The Acquisition of Physical Knowledge in Infancy: A Summary in
  Eight Lessons}, chapter~3, pp.\  47--83.
\newblock Wiley-Blackwell, 2007.
\newblock ISBN 9780470996652.
\newblock \doi{10.1002/9780470996652.ch3}.

\bibitem[Barto et~al.(2004)Barto, Singh, and
  Chentanez]{intrinsically_motivated_hierarcial_skills}
Barto, A.~G., Singh, S., and Chentanez, N.
\newblock Intrinsically motivated learning of hierarchical collections of
  skills.
\newblock In \emph{Proceedings of International Conference on Developmental
  Learning (ICDL)}. MIT Press, Cambridge, MA, 2004.

\bibitem[Battaglia et~al.(2016)Battaglia, Pascanu, Lai, Rezende, and
  kavukcuoglu]{InteratcionNetwork}
Battaglia, P., Pascanu, R., Lai, M., Rezende, D.~J., and kavukcuoglu, K.
\newblock Interaction networks for learning about objects, relations and
  physics.
\newblock In \emph{Proceedings of the 30th International Conference on Neural
  Information Processing Systems}, NIPS'16, pp.\  4509--4517, USA, 2016. Curran
  Associates Inc.
\newblock ISBN 978-1-5108-3881-9.

\bibitem[Battaglia et~al.(2013)Battaglia, Hamrick, and
  Tenenbaum]{Battaglia18327_simulation_as_an_engine}
Battaglia, P.~W., Hamrick, J.~B., and Tenenbaum, J.~B.
\newblock Simulation as an engine of physical scene understanding.
\newblock \emph{Proceedings of the National Academy of Sciences}, 110\penalty0
  (45):\penalty0 18327--18332, 2013.
\newblock ISSN 0027-8424.
\newblock \doi{10.1073/pnas.1306572110}.

\bibitem[Berlyne(1966)]{Berlyne25_curiosity_and_exploration}
Berlyne, D.~E.
\newblock Curiosity and exploration.
\newblock \emph{Science}, 153\penalty0 (3731):\penalty0 25--33, 1966.
\newblock ISSN 0036-8075.
\newblock \doi{10.1126/science.153.3731.25}.

\bibitem[Burda et~al.(2019)Burda, Edwards, Pathak, Storkey, Darrell, and
  Efros]{pathak18largescale}
Burda, Y., Edwards, H., Pathak, D., Storkey, A., Darrell, T., and Efros, A.~A.
\newblock Large-scale study of curiosity-driven learning.
\newblock In \emph{ICLR}, 2019.

\bibitem[Chang et~al.(2016)Chang, Ullman, Torralba, and
  Tenenbaum]{chang2016compositional}
Chang, M.~B., Ullman, T., Torralba, A., and Tenenbaum, J.~B.
\newblock A compositional object-based approach to learning physical dynamics.
\newblock \emph{arXiv preprint arXiv:1612.00341}, 2016.

\bibitem[Chentanez et~al.(2005)Chentanez, Barto, and Singh]{intrinsic_rl}
Chentanez, N., Barto, A.~G., and Singh, S.~P.
\newblock Intrinsically motivated reinforcement learning.
\newblock In Saul, L.~K., Weiss, Y., and Bottou, L. (eds.), \emph{Advances in
  Neural Information Processing Systems 17}, pp.\  1281--1288. MIT Press, 2005.

\bibitem[Diuk et~al.(2008)Diuk, Cohen, and Littman]{Diuk_ORE}
Diuk, C., Cohen, A., and Littman, M.~L.
\newblock An object-oriented representation for efficient reinforcement
  learning.
\newblock In \emph{Proceedings of the 25th International Conference on Machine
  Learning}, ICML '08, pp.\  240--247, New York, NY, USA, 2008. ACM.
\newblock ISBN 978-1-60558-205-4.
\newblock \doi{10.1145/1390156.1390187}.

\bibitem[Fragkiadaki et~al.(2015)Fragkiadaki, Agrawal, Levine, and
  Malik]{Fragkiadaki_billiards}
Fragkiadaki, K., Agrawal, P., Levine, S., and Malik, J.
\newblock Learning visual predictive models of physics for playing billiards.
\newblock \emph{CoRR}, abs/1511.07404, 2015.
\newblock URL \url{http://arxiv.org/abs/1511.07404}.

\bibitem[Hamrick(2011)]{internal_physics_models}
Hamrick, J.
\newblock Internal physics models guide probabilistic judgments about object
  dynamics.
\newblock 01 2011.

\bibitem[Lake et~al.(2017)Lake, Ullman, Tenenbaum, and
  Gershman]{lake_ullman_tenenbaum_gershman_2017}
Lake, B.~M., Ullman, T.~D., Tenenbaum, J.~B., and Gershman, S.~J.
\newblock Building machines that learn and think like people.
\newblock \emph{Behavioral and Brain Sciences}, 40:\penalty0 e253, 2017.

\bibitem[Lerer et~al.(2016)Lerer, Gross, and Fergus]{blocktowers}
Lerer, A., Gross, S., and Fergus, R.
\newblock Learning physical intuition of block towers by example.
\newblock In Balcan, M.~F. and Weinberger, K.~Q. (eds.), \emph{Proceedings of
  The 33rd International Conference on Machine Learning}, volume~48 of
  \emph{Proceedings of Machine Learning Research}, pp.\  430--438, New York,
  New York, USA, 20--22 Jun 2016. PMLR.

\bibitem[Mnih et~al.(2015)Mnih, Kavukcuoglu, Silver, Rusu, Veness, Bellemare,
  Graves, Riedmiller, Fidjeland, Ostrovski, Petersen, Beattie, Sadik,
  Antonoglou, King, Kumaran, Wierstra, Legg, and Hassabis]{mnih-dqn-2015}
Mnih, V., Kavukcuoglu, K., Silver, D., Rusu, A.~A., Veness, J., Bellemare,
  M.~G., Graves, A., Riedmiller, M., Fidjeland, A.~K., Ostrovski, G., Petersen,
  S., Beattie, C., Sadik, A., Antonoglou, I., King, H., Kumaran, D., Wierstra,
  D., Legg, S., and Hassabis, D.
\newblock Human-level control through deep reinforcement learning.
\newblock \emph{Nature}, 518\penalty0 (7540):\penalty0 529--533, 02 2015.

\bibitem[Oh et~al.(2017)Oh, Singh, and Lee]{value_prediction_network}
Oh, J., Singh, S., and Lee, H.
\newblock Value prediction network.
\newblock In Guyon, I., Luxburg, U.~V., Bengio, S., Wallach, H., Fergus, R.,
  Vishwanathan, S., and Garnett, R. (eds.), \emph{Advances in Neural
  Information Processing Systems 30}. 2017.

\bibitem[Pathak et~al.(2017)Pathak, Agrawal, Efros, and
  Darrell]{pathakICMl17curiosity}
Pathak, D., Agrawal, P., Efros, A.~A., and Darrell, T.
\newblock Curiosity-driven exploration by self-supervised prediction.
\newblock In \emph{ICML}, 2017.

\bibitem[Schaul et~al.(2015)Schaul, Quan, Antonoglou, and
  Silver]{prioritizedRB}
Schaul, T., Quan, J., Antonoglou, I., and Silver, D.
\newblock Prioritized experience replay.
\newblock \emph{CoRR}, abs/1511.05952, 2015.

\bibitem[Schmidhuber(1991)]{Schmidhuber91apossibility_for_curiosity_boredom}
Schmidhuber, J.
\newblock A possibility for implementing curiosity and boredom in
  model-building neural controllers, 1991.

\bibitem[Schmidhuber(2010)]{Formal_theory_of_creativity_fun_im}
Schmidhuber, J.
\newblock Formal theory of creativity, fun, and intrinsic motivation
  (1990–2010).
\newblock \emph{IEEE Transactions on Autonomous Mental Development}, 2\penalty0
  (3):\penalty0 230--247, Sept 2010.
\newblock ISSN 1943-0604.
\newblock \doi{10.1109/TAMD.2010.2056368}.

\bibitem[Schulman et~al.(2015)Schulman, Levine, Moritz, Jordan, and
  Abbeel]{Schulman:2015:TRP:3045118.3045319}
Schulman, J., Levine, S., Moritz, P., Jordan, M., and Abbeel, P.
\newblock Trust region policy optimization.
\newblock In \emph{Proceedings of the 32Nd International Conference on
  International Conference on Machine Learning - Volume 37}, ICML'15, pp.\
  1889--1897. JMLR.org, 2015.

\bibitem[Schulman et~al.(2017)Schulman, Wolski, Dhariwal, Radford, and
  Klimov]{DBLP:journals/corr/SchulmanWDRK17}
Schulman, J., Wolski, F., Dhariwal, P., Radford, A., and Klimov, O.
\newblock Proximal policy optimization algorithms.
\newblock \emph{CoRR}, abs/1707.06347, 2017.

\bibitem[Silver et~al.(2014)Silver, Lever, Heess, Degris, Wierstra, and
  Riedmiller]{Silver:2014:DPG:3044805.3044850}
Silver, D., Lever, G., Heess, N., Degris, T., Wierstra, D., and Riedmiller, M.
\newblock Deterministic policy gradient algorithms.
\newblock In \emph{Proceedings of the 31st International Conference on
  International Conference on Machine Learning - Volume 32}, ICML'14, pp.\
  I--387--I--395. JMLR.org, 2014.

\bibitem[Stahl \& Feigenson(2015)Stahl and Feigenson]{Stahl91}
Stahl, A.~E. and Feigenson, L.
\newblock Observing the unexpected enhances infants{\textquoteright} learning
  and exploration.
\newblock \emph{Science}, 348\penalty0 (6230):\penalty0 91--94, 2015.
\newblock ISSN 0036-8075.
\newblock \doi{10.1126/science.aaa3799}.
\newblock URL \url{http://science.sciencemag.org/content/348/6230/91}.

\bibitem[Stahl \& Feigenson(2017)Stahl and Feigenson]{STAHL20171}
Stahl, A.~E. and Feigenson, L.
\newblock Expectancy violations promote learning in young children.
\newblock \emph{Cognition}, 163:\penalty0 1 -- 14, 2017.
\newblock ISSN 0010-0277.
\newblock \doi{https://doi.org/10.1016/j.cognition.2017.02.008}.
\newblock URL
  \url{http://www.sciencedirect.com/science/article/pii/S0010027717300380}.

\bibitem[Sutton(1990)]{Sutton90integratedarchitectures}
Sutton, R.~S.
\newblock Integrated architectures for learning, planning, and reacting based
  on approximating dynamic programming.
\newblock In \emph{In Proceedings of the Seventh International Conference on
  Machine Learning}, pp.\  216--224. Morgan Kaufmann, 1990.

\bibitem[Sutton \& Barto(1998)Sutton and Barto]{Sutton:1998:intro_to_rl}
Sutton, R.~S. and Barto, A.~G.
\newblock \emph{Introduction to Reinforcement Learning}.
\newblock MIT Press, Cambridge, MA, USA, 1st edition, 1998.
\newblock ISBN 0262193981.

\bibitem[Watters et~al.(2017)Watters, Zoran, Weber, Battaglia, Pascanu, and
  Tacchetti]{visual_interaction_networks}
Watters, N., Zoran, D., Weber, T., Battaglia, P., Pascanu, R., and Tacchetti,
  A.
\newblock Visual interaction networks: Learning a physics simulator from video.
\newblock In Guyon, I., Luxburg, U.~V., Bengio, S., Wallach, H., Fergus, R.,
  Vishwanathan, S., and Garnett, R. (eds.), \emph{Advances in Neural
  Information Processing Systems 30}, pp.\  4539--4547. Curran Associates,
  Inc., 2017.

\end{thebibliography}
\bibliographystyle{icml2019}


\end{document}